\pdfoutput=1

\documentclass[11pt]{article}

\usepackage[final]{acl}
\usepackage{amsmath}
\usepackage{amsfonts,amssymb}
\usepackage{times}
\usepackage{CJK}
\usepackage{latexsym}
\usepackage{subfiles}
\usepackage{booktabs}
\usepackage{multirow}
\usepackage{array,tabularx,caption,boldline}
\usepackage{arydshln}
\usepackage{amsmath}
\usepackage{hyperref}
\DeclareMathOperator*{\argmax}{arg\,max}

\definecolor{DeepYellow}{rgb}{0.9, 0.7, 0.0} 

\usepackage{times}
\usepackage{latexsym}

\usepackage[T1]{fontenc}

\usepackage[utf8]{inputenc}

\usepackage{microtype}

\usepackage{inconsolata}

\usepackage{graphicx}

\newcommand{\ignore}[1]{}

\title{Improving LLM-based Document-level Machine Translation with Multi-Knowledge Fusion}

\author{
 \textbf{Bin Liu\textsuperscript{1}},
 \textbf{Xinglin Lyu\textsuperscript{2}},
 \textbf{Junhui Li\textsuperscript{1}},
 \textbf{Daimeng Wei\textsuperscript{2}},
\\
 \textbf{Min Zhang\textsuperscript{2}},
 \textbf{Shimin Tao\textsuperscript{2}},
 \textbf{Hao Yang\textsuperscript{2}}
\\
 \textsuperscript{1}School of Computer Science and Technology, Soochow University, Suzhou, China
 \\
 \textsuperscript{2}Huawei Translation Services Center, Beijing, China
\\
\texttt{20235227033@stu.suda.edu.cn,lijunhui@suda.edu.cn}
\\
\texttt{\{lvxinglin1,weidaimeng,zhangmin186,taoshimin,yanghao30\}@huawei.com}
}

\begin{document}
\maketitle
\begin{abstract}
Recent studies in prompting large language model (LLM) for document-level machine translation (DMT) primarily focus on the inter-sentence context by flatting the source document into a long sequence. This approach relies solely on the sequence of sentences within the document. However, the complexity of document-level sequences is greater than that of shorter sentence-level sequences, which may limit LLM's ability in DMT when only this single-source knowledge is used. In this paper, we propose an enhanced approach by incorporating multiple sources of knowledge, including both the document summarization and entity translation, to enhance the performance of LLM-based DMT. Given a source document, we first obtain its summarization and translation of entities via LLM as the additional knowledge. We then utilize LLMs to generate two translations of the source document by fusing these two single knowledge sources, respectively. Finally,  recognizing that different sources of knowledge may aid or hinder the translation of different sentences, we refine and rank the translations by leveraging a multi-knowledge fusion strategy to ensure the best results. Experimental results in eight document-level translation tasks show that our approach achieves an average improvement of 0.8, 0.6, and 0.4 COMET scores over the baseline without extra knowledge for \texttt{LLaMA3-8B-Instruct}, \texttt{Mistral-Nemo-Instruct}, and \texttt{GPT-4o-mini}, respectively. We will release our code on \url{https://github.com/gfvmjcdfg/Mutil_knowledge_fusion}.

\end{abstract}

\section{Introduction}
Large language model (LLM) has shown impressive performance across various natural language processing (NLP) tasks \cite{addams-etal-2023-from,dong-etal-2024-selfcollaboration}. Many researchers also explore how to utilize LLM to solve the document-level machine translation (DMT), in which LLM needs to capture the inter-sentence dependency for addressing discourse issues, such as pronoun translation and word translation inconsistency. Existing approaches can be roughly categorized into 1) supervised fine-tuning (SFT) approaches \cite{lyu-etal-2024-dempt,li-etal-2024-enhancing,wu-etal-2024-adapting} and 2) prompt engineering (PE) approaches \cite{wang-etal-2023-documentlevel,wu-hu-2023-exploring}. The former directly leverages the document-level parallel corpus to tune LLM via some parameter-efficient methods\ignore{, such as \citet {lyu-etal-2024-dempt} and \citet{li-etal-2024-enhancing}}, while the later mainly relies on the ability of LLM in in-context learning. \ignore{For example, \citet{wu-hu-2023-exploring} explore the effect of various prompts and prompting forms (single-turn or multi-turn) for the performance of LLM-base DMT. }Compared to SFT approaches, the PE approaches do not require additional computing resources to train or tune LLM, and are more resource-efficient. Therefore, in this paper we effectively explore LLM for DMT by a novel PE approach, called multi-knowledge fusion.

Despite its resource-efficient superiority, the context or knowledge utilized in the existing PE approach is limited. \ignore{For example, \citet{wang-etal-2023-documentlevel} translate the document via a sentence-by-sentence prompting form. When translating each sentence in a document, they only utilize the inter-sentence context/knowledge most related to the current sentence.}For example, \citet{wang-etal-2023-documentlevel} translate documents sentence by sentence, using only inter-sentence context/knowledge most relevant to the current sentence. For sentence-level machine translation, benefiting from the powerful in-context learning ability, randomly sampling bilingual parallel sentence pairs as prompts can effectively enhance the translation abilities of LLMs. However, different from sentence-level translation, the single knowledge fusion strategy may not effectively solve the tough discourse problem in DMT. Professional human translators typically explicit multi-knowledge information to ensure that its translation has stronger coherence and lexical cohesion when translating a source document, such the topics, keywords, and entity words. 

\begin{figure*}[t]
\centering
\includegraphics[width=\textwidth, trim={0cm 0cm 0cm 0cm}, clip]{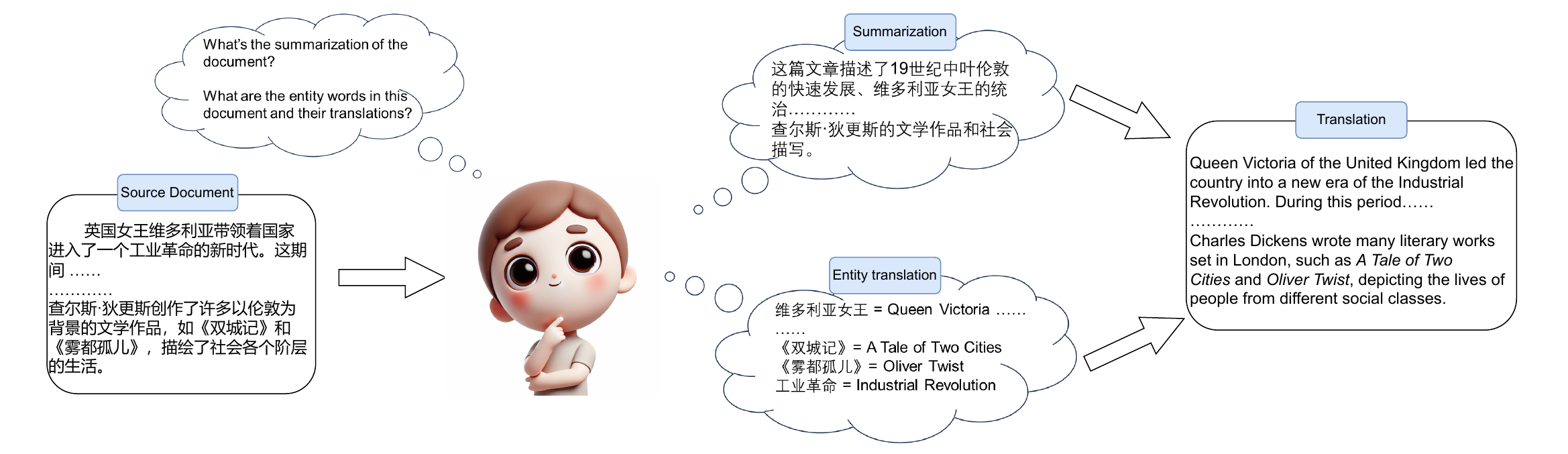}
\caption{Illustration of a professional translator translating a document from Chinese to English.}
\label{fig:illustration}
\end{figure*}

When translating a document, as depicted in Figure~\ref{fig:illustration}, a professional translator first reads the entire text to understand its content. Additionally, the translator may highlight key entities within the document and consider their translations in advance. To mimic this behavior, we propose a multi-knowledge fusion approach to prompt LLMs to generate better document translation. Motivated by~\citet{he-etal-2024-exploring}, our approach consists of three essential steps: 
\begin{itemize}
\item \textit{Document-Level Knowledge Acquisition}: In this initial step, we prompt the LLM to extract two critical types of inter-sentence knowledge: \textbf{summarization} and \textbf{entity translation}. Summarization enables readers to quickly grasp the main ideas of lengthy documents, facilitating a better understanding of the key points. This improves comprehension of the overall context allows the LLM to produce more coherent translations. Additionally, knowledge of entity translation aids in organizing and structuring documents by maintaining consistency in the translation of entities, which enhances the clarity\ignore{ and navigability} of the output.
\ignore{Generally, the repeat entity words in the source document should be consistently translated into the same. The knowledge in entity words translation will guide the LLM to have a stronger consistency preference when translating the same entity words. While summarization can empower LLM to grasp the overall context of the translation, resulting in translation with stronger coherence.}
\item \textit{Single-Knowledge Integration}: This step involves incorporating specific pieces of acquired knowledge into the process of document-level machine translation. While not every sentence in the document will need this integration, certain sentences will benefit from the added knowledge.\ignore{By effectively utilizing each individual piece of knowledge, this approach improves the overall quality and reliability of the translation output.}


\item \textit{Multi-Knowledge Fusion}: In the final step, we \ignore{employ a post-editing technique to }re-evaluate and rank translations that integrate multiple facets of knowledge. This process involves merging various elements and refining the translations to ensure that the final output accurately and comprehensively represents the source document.

\end{itemize}

\ignore{
\textcolor{red}{To more effectively alleviate the complex discourse issues in DMT}, we propose a multi-knowledge fusion strategy to prompt LLMs to generate
better document translation. Our approach consist of three essential steps: Firstly, \textit{knowledge acquisition} step, where we prompt LLM to elicit two additional inter-sentence knowledge, \textbf{entity words translation} and \textbf{summarization}.\footnote{We also try to integrate other aspects of knowledge such as **. Please refer to **. } Generally, the repeat entity words in the source document should be consistently translated into the same. The knowledge in entity words translation will guide the LLM to have a stronger consistency preference when translating the same entity words. While summarization can empower LLM to grasp the overall context of the translation, resulting in translation with stronger coherence.
Secondly, \textit{single-knowledge fusion} step, where we independently integrate the knowledge of entity words translation and summarization, prompting LLM toward generating more consistent and coherent translations. Finally, \textit{multi-knowledge fusion} step, where we employ a post-edit method to integrate all aspects of knowledge, including the \textbf{basic knowledge},\footnote{Basic knowledge refers to the original information contained within the given document, i.e., the source document itself.} into the final translation based on reference-free quality estimation (QE). 
}

Overall, our main contributions in this work can be summarized as follows:

\begin{itemize}
    \item We introduce two additional aspects of knowledge, entity word translation and summarization, guiding LLMs to generate better document translations.
    \item Interestingly, we observe that a single type of knowledge can improve the translation of certain sentences in a document while potentially harming some others. To address this, we propose a novel multi-knowledge fusion strategy to enhance the performance of LLM-based DMT further.
    \item  Upon various LLMs, including \texttt{LLaMA3-8B\--Instruct}, \texttt{Mixtral-Nemo-Instruct} and \texttt{GPT-4o-mini}, we demonstrate the effectiveness of the proposed approach across eight document-level machine translation directions. And additional analysis further verify the superiority of the proposed approach in addressing discourse issues. 
\end{itemize}



\section{Background}
In this section, we briefly introduce the conventional DMT and prompting LLM for DMT.
\subsection{Conventional DMT}
Given a source document {\small $\mathcal{X} = \{{X_{1},\cdots, X_{N}}\}$} with $N$ sentences, the conventional DMT model maps each sentence {\small${X_i}=\{{x_1},\cdots,{x_{|X_i|}}$\}} with {\small $|X_i|$} words into the target sentence {\small $Y_i$} by leveraging the inter-sentence context {\small$C$}. More specifically, the target document {\small $\mathcal{Y}$} is generated as follows:
\begin{equation}
    \small
    \mathcal{Y} = \argmax P\left(\mathcal{Y}|\mathcal{X}; \theta\right),
\end{equation}
\begin{equation}
\small
P(\mathcal{Y}|\mathcal{X}; \theta) = \prod_{i=1}^{N}\prod_{j=1}^{|Y_i|} P(y_{j}^i|y_{<j}^{i}, X_i, C; \theta),
\end{equation}
where {\small $\theta$} denotes the model parameters and {\small $|Y_{i}|$} is the length of sentence {\small $Y_{i}$}. {\small$C$} includes both the source-side and target-side inter-sentence contexts.

\begin{figure*}[t]
\centering
\includegraphics[width=\textwidth, trim={1cm 1cm 1cm 0.8cm}, clip]{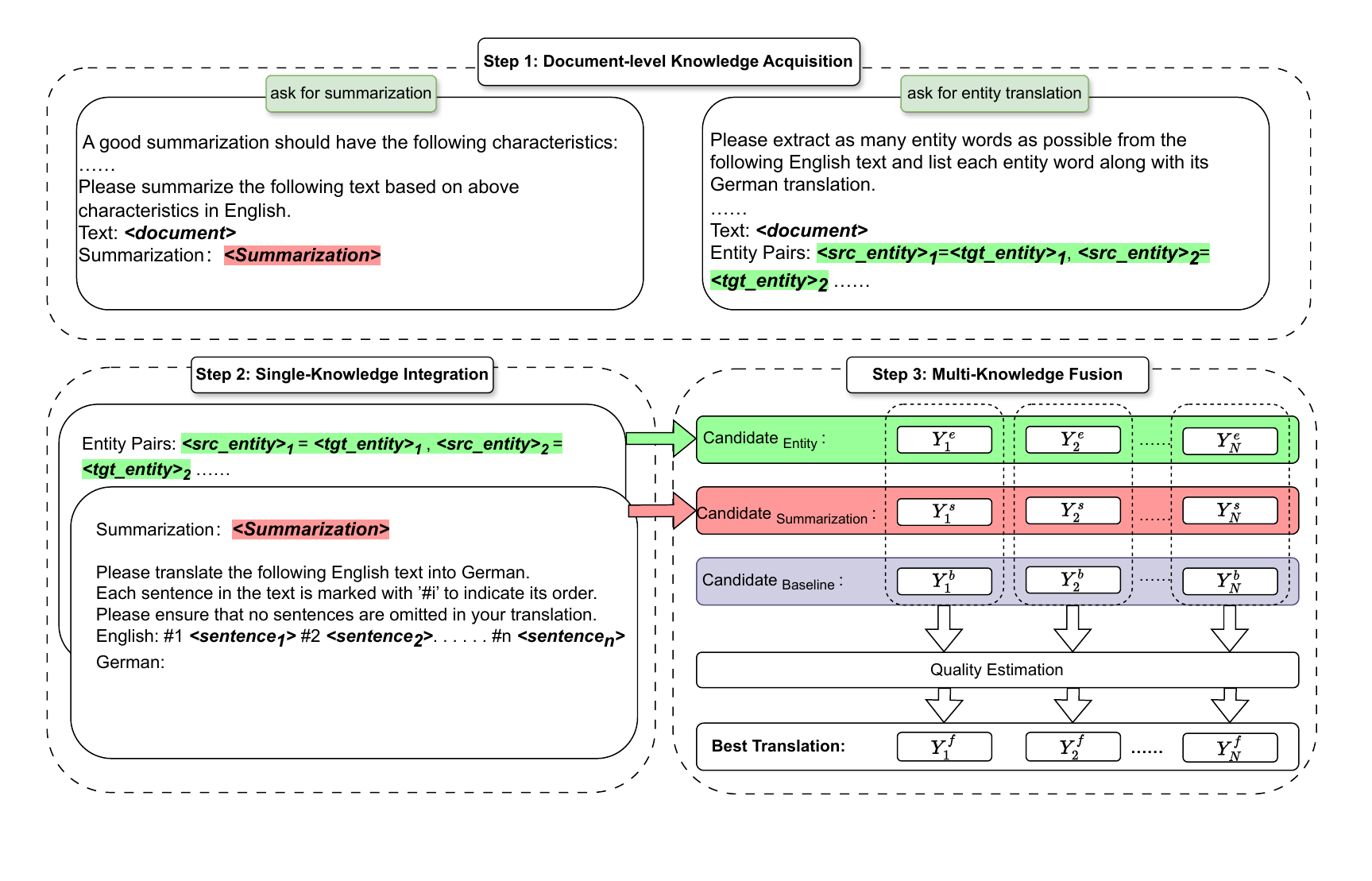}
\caption{Illustration of our approach, which mimics the human-preferring translating process. Given a document, we first obtain its summarization and entity translation ({\bf step 1}), then prompt LLMs to generate better document translation based on these additional knowledge ({\bf step 2} and {\bf step 3}).}
\label{fig:approach}
\end{figure*}

\subsection{Prompting LLM for DMT}
Different from the conventional DMT, the LLM has impressive ability in instruction following. We can prompt LLM to translate the given document into the target document via concatenating a translation instruction. Similarly, LLM generates the translation of a given document as follows:

\begin{equation}
\label{equ:llmmt-1}
\small
\mathcal{Y} = \argmax P(\mathcal{Y}|\mathcal{X},\mathcal{P}; \theta),
\end{equation}
\begin{equation}
\label{equ:llmmt-2}
\small
P\left(\mathcal{Y}|\mathcal{X},\mathcal{P};\theta\right) = \prod_{i=1}^{N}\prod_{j=1}^{|Y_i|} P(y_{j}^i|y_{<j}^{i}, X_i, C, \mathcal{P}; \theta),
\end{equation}
where {\small $\mathcal{P}$} is the instruction text.\ignore{ Previous efforts in prompting LLM for DMT mainly focus on the design of {\small $\mathcal{P}$}.}

\section{Methodology}
\label{Methodology}

\ignore{In this section, we introduce our proposed multi-knowledge fusion approach for LLM-based DMT in details. }As shown in Figure \ref{fig:approach}, our approach consists of three essential steps: document-level knowledge acquisition, single-knowledge integration, and multi-knowledge fusion.

\begin{table*}[!h]
\centering
\small
\resizebox{\textwidth}{!}{
\begin{tabular}{lp{0.25\textwidth}p{0.65\textwidth}} 
\toprule
\textbf{ID} & \textbf{Task} & \textbf{Prompt Template} \\
\midrule
\#1 & Summarization \newline Acquisition & A good summarization should have the following characteristics: \newline - Include the main points \newline - Include key details \newline - Be concise (no more than 3 sentences) \newline - Remain objective \newline Please summarize the following text based on above characteristics in English. \newline Text: \textbf{\textit{<document>}} \newline Summarization: \\
\midrule

\#2 & Entity Translation \newline Acquisition& Please extract as many entity words as possible from the following \textbf{\textit{\textless{}src\_lang\textgreater{}}} text and list each entity word along with its \textbf{\textit{\textless{}tgt\_lang\textgreater{}}} translation. Entity words include but are not limited to: Person, Organization, Location, Date, Money, Work of Art, Product, Event, Occupation, Social Group, Animal, and so on. \newline Text: \textbf{\textit{\textless{}document\textgreater{}}} \newline Entity Pairs: \\
\midrule

\#3 & Prompting LLM for DMT w/o Knowledge & Please translate the following \textbf{\textit{\textless{}src\_lang\textgreater{}}} text into \textbf{\textit{\textless{}tgt\_lang\textgreater{}}}. Each sentence in the text is marked with '\#i' to indicate its order. Please ensure that no sentences are omitted in your translation. \newline \textbf{\textit{\textless{}src\_lang\textgreater{}}}: \#1 \textbf{\textit{\textless{}sentence\textsubscript{1}\textgreater{}}} \#2 \textbf{\textit{\textless{}sentence\textsubscript{2}\textgreater{}}} ……  \ignore{\#n \textbf{\textit{\textless{}sentence\textsubscript{n}\textgreater{}}}} 
  \newline \textbf{\textit{\textless{}tgt\_lang\textgreater{}}}: \\
\midrule

\#4 & Prompting LLM for DMT with Summarization & Summarization: \textbf{\textit{\textless{}summarization\textgreater{}}} \newline \newline Please translate the following \textbf{\textit{\textless{}src\_lang\textgreater{}}} text into \textbf{\textit{\textless{}tgt\_lang\textgreater{}}}. Each sentence in the text is marked with '\#i' to indicate its order. Please ensure that no sentences are omitted in your translation. \newline \textbf{\textit{\textless{}src\_lang\textgreater{}}}: \#1 \textbf{\textit{\textless{}sentence\textsubscript{1}\textgreater{}}} \#2 \textbf{\textit{\textless{}sentence\textsubscript{2}\textgreater{}}} …… \ignore{\#n \textbf{\textit{\textless{}sentence\textsubscript{n}\textgreater{}}}} \newline \textbf{\textit{\textless{}tgt\_lang\textgreater{}}}: \\
\midrule

\#5 & Prompting LLM for DMT with Entity Translation & Entity pairs: \textbf{\textit{\textless{}src\textsubscript{E\textsubscript{1}}\textgreater{}}} = \textbf{\textit{\textless{}tgt\textsubscript{E\textsubscript{1}}\textgreater{}}} , \textbf{\textit{\textless{}src\textsubscript{E\textsubscript{2}}\textgreater{}}} = \textbf{\textit{\textless{}tgt\textsubscript{E\textsubscript{2}}\textgreater{}}} , …… \ignore{\textbf{\textit{\textless{}src\textsubscript{E\textsubscript{n}}\textgreater{}}} = \textbf{\textit{\textless{}tgt\textsubscript{E\textsubscript{n}}\textgreater{}}}}
 \newline \newline Please translate the following \textbf{\textit{\textless{}src\_lang\textgreater{}}} text into \textbf{\textit{\textless{}tgt\_lang\textgreater{}}}. Each sentence in the text is marked with `\#i' to indicate its order. Please ensure that no sentences are omitted in your translation. \newline \textbf{\textit{\textless{}src\_lang\textgreater{}}}: \#1 \textbf{\textit{\textless{}sentence\textsubscript{1}\textgreater{}}} \#2 \textbf{\textit{\textless{}sentence\textsubscript{2}\textgreater{}}} …… \ignore{\#n \textbf{\textit{\textless{}sentence\textsubscript{n}\textgreater{}}}}  \newline \textbf{\textit{\textless{}tgt\_lang\textgreater{}}}: \\
\bottomrule
\end{tabular}
}
\caption{Prompt templates used for document-level knowledge acquisition (\#1, \#2 and \#3) and single-knowledge integration (\#4 and \#5).}
\label{tbl:main_prompt}
\end{table*}

\subsection{Document-Level Knowledge Acquisition}
Given the source document {\small$\mathcal{X}$}, we use \textbf{summarization } and \textbf{entity translation } to provide additional context, helping the LLM produce a more accurate and fluent translation of the document.

\paragraph{Summarization Knowledge.} A summarization provides an overall view of the document, capturing its main themes and key points. By generating a summarization, the LLM gains a clearer understanding of the document's overall context, which helps in translating complex ideas, metaphors, and culturally specific content more accurately. Research by \citet{pu-etal-2023-summarization} and \citet{tianyi-etal-2024-benchmarking} demonstrate that large models often produce summarization with superior fluency and authenticity compared to humans. Thus, we first prompt the LLM to summarize the source document, with the specific prompt outlined in row \#1 of Table \ref{tbl:main_prompt}.

\paragraph{Entity Translation Knowledge.} Entity translation knowledge can enhance the translation consistency of specific terms in the document \cite{lyu-etal-2021-encouraging}. Additionally, identifying entities in the text can reduce the occurrence of untranslated segments. The entities used here include not only the general types of entities such as \textit{name, place} and \textit{organization} but also \textit{events}. We prompt the LLM to fetch the entity translation knowledge with the instruction shown in row \#2 of Table \ref{tbl:main_prompt}.

\subsection{Single-Knowledge Integration}

As long as we obtain the extracted knowledge from the given document, i.e., the summarization or entity translation, we explicitly integrate the \ignore{summarization or entity translation }knowledge into $\mathcal{P}$, as shown in rows \#4 and \#5 of Table~\ref{tbl:main_prompt}, prompting LLM to generate more accurate translation by Eq. \ref{equ:llmmt-1} and \ref{equ:llmmt-2}.

After that, \ignore{we obtain two different translations of the given source document. We denote {\small$\mathcal{Y}^{s}$} and {\small$\mathcal{Y}^{e}$} as translations with consideration of the summarization and entity translation knowledge, respectively. Expect {\small$\mathcal{Y}^{s}$} and {\small$\mathcal{Y}^{e}$}, we elicit another translation {\small$\mathcal{Y}^b$} that do not explicitly consider any additional knowledge, as shown in row \#3 of Table~\ref{tbl:main_prompt}.}we generate two different translations of the source document: {\small$\mathcal{Y}^{s}$}, incorporating summarization knowledge, and {\small$\mathcal{Y}^{e}$}, incorporating entity translation knowledge. Additionally, we produce {\small$\mathcal{Y}^b$}, a baseline translation without additional knowledge, as detailed in row \#3 of Table~\ref{tbl:main_prompt}.

\subsection{Multi-Knowledge Fusion}
\label{sec:multi-fusion}
Intuitively, each piece of knowledge does not always be beneficial to the translations of all sentences within the source document. The summarization knowledge can promote the translation quality of the sentences that are more related to main topic of the document. While the translation quality of sentences in which appear more entity words tend to be benefit from the entity translation knowledge. To better leverage these different types of knowledge, we fuse different knowledge to obtain a better translation by integrating {\small$\mathcal{Y}^{s}$}, {\small$\mathcal{Y}^{e}$} and {\small$\mathcal{Y}^{b}$}. Specifically, we assume each translation of {\small$\mathcal{X}$}, i.e., {\small$\mathcal{Y}^{s}= \{Y_1^s,\cdots,Y_N^s\}$}, {\small$\mathcal{Y}^{e}= \{Y_1^e,\cdots,Y_N^e\}$} and {\small$\mathcal{Y}^{b}= \{Y_1^b,\cdots,Y_N^b\}$}, contain $N$ translation segments, corresponding to the translations of {\small$N$} sentence within {\small$\mathcal{X}$}.\footnote{Sentence-level translations can be readily obtained using the instructions provided in row \#3 of Table~\ref{tbl:format_prompt} in Appendix~\ref{apdx:prompt_format}. In row \#1 and \#2 ,we can see our instructions for obtaining stable format of summarization and entity translation.} We first select its best translation, {\small $Y_i^f$}, for each sentence {\small $X_i$} in {\small$\mathcal{X}$} from {\small$Y_i^{s}$}, {\small$Y_i^{e}$} and {\small$Y_i^{b}$}:

\begin{equation}
\small
\label{equ:scoring}
Y_i^f = \argmax~S(Y, X_i), 
\end{equation}
where {\small $Y\in\{Y_i^s, Y_i^e, Y_i^b\}$} and {\small$S(\cdot)$} is a reference-free scoring function. Then the final translation of {\small$\mathcal{X}$} can be formulated as {\small $\mathcal{Y}^f = \{Y_1^f,\cdots,Y_N^f\}$}.

\section{Experimentation}
\begin{table*}[th]
\centering
\small 
\begin{tabular}{lccccccccc}
\toprule
\textbf{System} & \textbf{En$\rightarrow$De} & \textbf{De$\rightarrow$En} & \textbf{En$\rightarrow$Es} & \textbf{Es$\rightarrow$En} & \textbf{En$\rightarrow$Ru} & \textbf{Ru$\rightarrow$En} & \textbf{En$\rightarrow$Fr} & \textbf{Fr$\rightarrow$En} & \textbf{Average}\\ 
\midrule
\multicolumn{9}{c}{\texttt{LLaMA3-8B-Instruct}} \\ 
\midrule
Baseline     & 85.2  & 88.2  & 87.1  & 88.8  & 83.8  & 83.9  & 84.9  & 87.0  &86.1\\
Reranking    & 85.7  & 88.4  & 87.4  & 88.9  & 84.5  & 84.2  & 85.3  & 87.2  &86.5\\ \hdashline
SuMT         & 85.3  & 88.3  & 87.2  & 88.8  & 83.7  & 84.1  & 85.0  & 87.3  &86.2\\ 
EnMT         & 85.3  & 88.3  & 86.9  & 88.4  & 83.4  & 83.9  & 84.8  & 86.9  &86.0\\ \hdashline
KFMT         & 86.1  & 88.6  & 87.8  & 89.0  & 85.5  & 84.7  & 85.8  & 87.6  &86.9\\
KFMT$_{\mathbf{Oracle}}$ & \textbf{86.4} & \textbf{88.8} & \textbf{88.2} & \textbf{89.4} & \textbf{86.0} & \textbf{85.1} & \textbf{86.3} & \textbf{88.0} &\textbf{87.3}\\ 
\midrule
\multicolumn{9}{c}{\texttt{Mistral-Nemo-Instruct}} \\ 
\midrule
Baseline     & 86.5  & 89.0  & 87.3  & 89.4  & 87.0  & 85.2  & 85.9  & 88.0  &87.3\\ 
Reranking    & 87.1  & 89.0  & 87.7  & 89.4  & 87.4  & 85.2  & 86.3  & 88.0  &87.5\\ \hdashline
SuMT         & 86.5  & 88.5  & 87.5  & 89.1  & 87.0  & 84.7  & 86.0  & 87.6  &87.1\\ 
EnMT         & 86.6  & 88.8  & 87.3  & 89.2  & 87.0  & 85.0  & 86.0  & 87.8  &87.2\\ \hdashline
KFMT         & 87.6  & 89.3  & 88.2  & 89.6  & 87.9  & 85.4  & 86.7  & 88.1  &87.9\\
KFMT$_{\mathbf{Oracle}}$ & \textbf{88.1} & \textbf{89.6} & \textbf{88.5} & \textbf{89.9} & \textbf{88.3} & \textbf{85.9} & \textbf{87.1} & \textbf{88.6} &\textbf{88.3}\\ 
\midrule
\multicolumn{9}{c}{\texttt{GPT-4o-mini}} \\ 
\midrule
Baseline     & 88.5  & 89.3  & 88.9  & 89.6  & 88.7  & 85.5  & 87.3  & 88.1  &88.2\\ 
Reranking    & 88.7  & 89.4  & 89.1  & 89.7  & 88.8  & 85.6  & 87.5  & 88.2  &88.4\\ \hdashline
SuMT         & 88.5  & 89.3  & 88.9  & 89.6  & 88.7  & 85.5  & 87.3  & 88.1  &88.2\\ 
EnMT         & 88.2  & 89.1  & 88.7  & 89.4  & 88.3  & 85.3  & 87.0  & 87.9  &88.0\\ \hdashline
KFMT         & 88.9  & 89.6  & 89.3  & 89.9  & 89.2  & 85.8  & 87.7  & 88.4  &88.6\\
KFMT$_{\mathbf{Oracle}}$ & \textbf{89.1} & \textbf{89.8} & \textbf{89.4} & \textbf{90.0} & \textbf{89.4} & \textbf{86.1} & \textbf{87.8} & \textbf{88.6} &\textbf{88.8}\\ 
\bottomrule
\end{tabular}
\caption{Performance in reference-based COMET score. {\bf Baseline} shows results from prompting LLMs without additional knowledge. {\bf Reranking} denotes ensemble results by reranking three translations generated by {\bf Baseline}.  {\bf SuMT} and {\bf EnMT} are the results for prompting LLMs via \textit{summarization} and \textit{entity translation} knowledge, respectively. {\bf KFMT} and {\bf $\text{KFMT}_\text{Oracle}$} are the results with the multi-knowledge fusion strategy, where {\bf KFMT} uses a reference-free scoring function and {\bf $\text{KFMT}_\text{Oracle}$} uses a reference-based one to select the best translation.}
\label{tbl:comet_res}
\end{table*}

We verify the effectiveness of our approach on three popular LLMs, including open-source and closed-source, across eight translation directions.
\subsection{Experimental Settings}
\paragraph{LLMs and Datasets.} We evaluate our approach upon three LLMs, including \texttt{GPT-4o-mini}
\cite{openai-etal-gpt4o-2024}\footnote{\url{https://openai.com/research/gpt-4}}, \texttt{LLaMA3-8B-Instruct} \cite{metaai-etal-llama3-2024}\footnote{\url{https://ai.meta.com/blog/meta-llama-3}} and \texttt{Mistral-Nemo-Instruct} \cite{mistralai-etal-mistral-2024}\footnote{\url{https://mistral.ai/news/mistral-nemo/}}. Our test set are extracted from WMT 2023 News Commentary v18, including  English (En) $\rightleftharpoons$ \{German (De), French (Fr), Spanish (Es), Russian (Ru)\} eight translation directions. The test set for each translation direction contains 150 document pairs. For additional details, please refer to Table~\ref{tbl:data_stat} in Appendix~\ref{apdx:data_stat}.

\paragraph{Inference Settings.} We run the inference of open-source LLMs, i.e., \texttt{LLaMA3-8B-Instruct} and \texttt{Mistral-Nemo-Instruct}, on a single NVIDIA V100 32GB GPU using greedy decoding strategy. For closed-source \texttt{GPT-4o-mini}, we run the inference by calling the official API. The temperature is set to 0 in the inference of all LLMs. In multi-knowledge fusion, we employ reference-free COMET\footnote{\url{https://huggingface.co/Unbabel/wmt22-cometkiwi-da}} as the scoring function in Eq. \ref{equ:scoring}.

\paragraph{Evaluation Metrics.} Following recent studies~ \cite{wang-etal-2023-documentlevel,li-etal-2024-enhancing,wu-etal-2024-adapting}, we report reference-based COMET \cite{rei-etal-2022-comet22} to evaluate system performance. Specifically, We use \texttt{wmt22-comet-da}\footnote{\url{https://huggingface.co/Unbabel/wmt22-comet-da}} as our evaluation model. For translation performance in dCOMET and BLEU, please refer to Table~\ref{tbl:dCOMET} and Table~\ref{tbl:bleu} in Appendix~\ref{apdx:bleuAndDComet}. Additionally, we report performance using the BlonDe metric \cite{jiang-etal-2022-blonde}, which evaluates discourse phenomena based on a set of discourse-related features, with further details available in Appendix~\ref{apdx:blonde}.

\subsection{Experimental Results}

For better demonstrating the effect of integrating knowledge, we also build a \texttt{Reranking} system that ranks three different translation generated by \texttt{Baseline}. Table \ref{tbl:comet_res} presents the main experimental results, which highlight the following observations:

\begin{itemize}
    \item Our multi-knowledge-fusion approach significantly enhances LLM performance on DMT. Specifically, our \texttt{KFMT} achieves an average improvement of 0.8, 0.6, and 0.4 COMET scores over the \texttt{Baseline} for \texttt{LLaMA3-\-8B-\-Instruct}, \texttt{Mistral-\-Nemo\--Instruct}, and \texttt{GPT\--4o\--mini}, respectively. \texttt{KFMT\textsubscript{Oracle}} represents the upper bound of our approach, with an average maximum improvement of 1.2, 1.0, 0.6 COMET scores across the three LLMs.

    \item Due to the predominance of English data in the training of LLMs, our proposed approach shows a more pronounced improvement in En$\rightarrow$\textit{X} translation tasks compared to \textit{X}$\rightarrow$En translation tasks. For instance, with \texttt{LLaMA3-8B-Instruct}, the average improvement for En$\rightarrow$\textit{X} translation tasks is 1.1, which is notably higher than the 0.5 improvement observed for \textit{X}$\rightarrow$En translation tasks.
    \item Our approach consistently outperforms the naive \texttt{Reranking} approach, which does not incorporate any additional knowledge during reranking. This further suggests the necessity of integrating diverse knowledge sources.
    \item The single-knowledge fusion methods, namely \texttt{SuMT} and \texttt{EnMT}, do not always show improvements over the \texttt{Baseline}. This indicates that the benefits of different types of knowledge are most pronounced in translations of sentences closely related to that specific knowledge, rather than in all sentences within a document. 
\end{itemize}

\subsection{Experimental Analysis}
\label{sec:expr_analysis}

To clarify the proposed approach, we conduct an in-depth analysis using the En$\rightleftharpoons$Ru and En$\rightleftharpoons$FR translation tasks as representatives. These analyses are carried out with the \texttt{LLaMA3-8B-Instruct} model to gain further insights. Additionally, Appendix~\ref{apdx:summarziation_entity} includes an analysis of summarization and entity translation accuracy.

\begin{figure}[!t]
\centering
\resizebox{0.99\columnwidth}{!}{
\includegraphics[trim={0cm 0cm 0cm 0cm}]{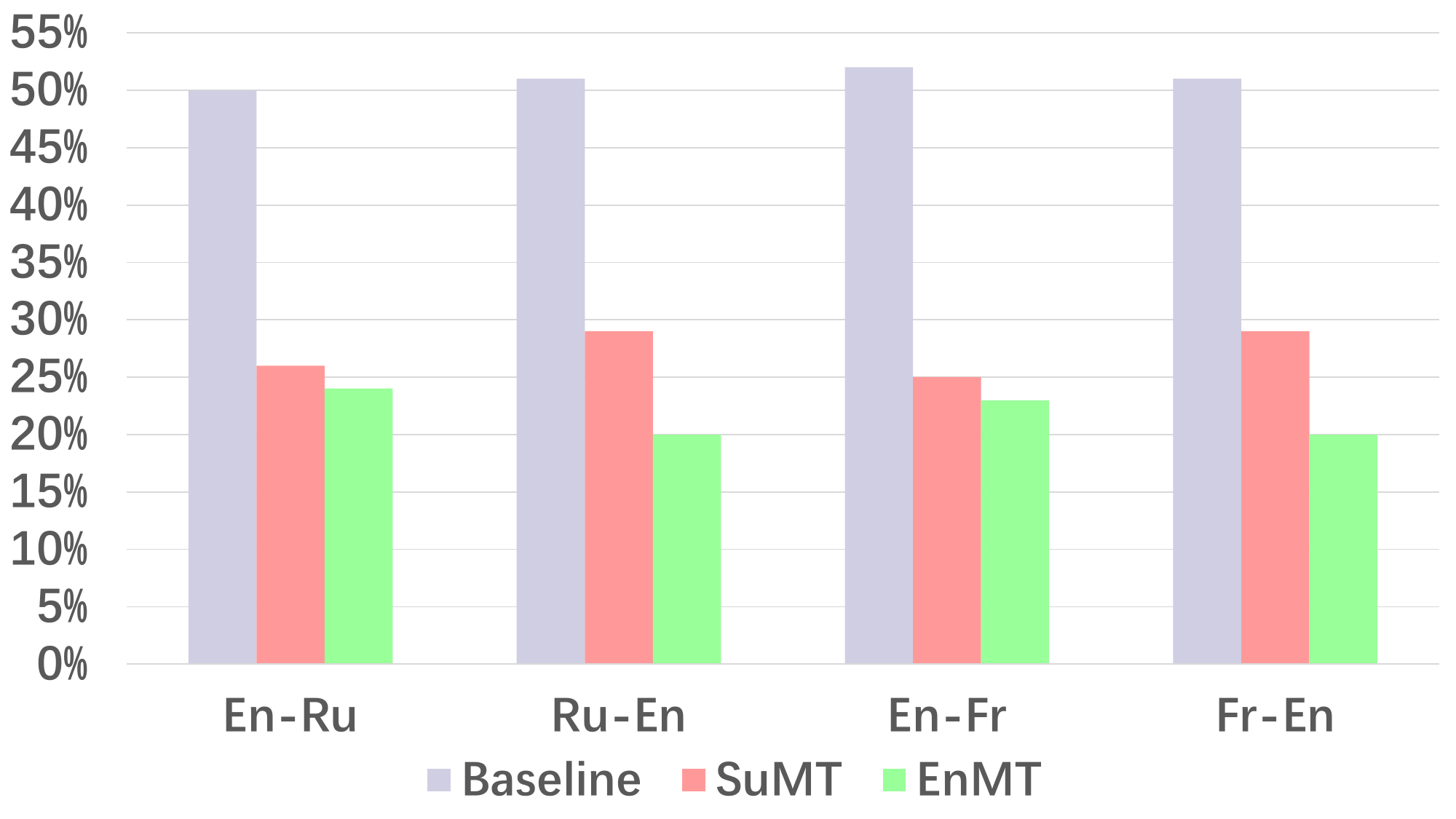}}
\caption{Visualization of the proportions of translations produced by the Baseline, SuMT, and EnMT systems relative to the total number of translations.}
\label{fig:propotion}
\end{figure}

\begin{table}[!t]
\centering
\small
\begin{tabular}{lcccc}
\toprule
\textbf{System}      & \textbf{En$\rightarrow$Ru} & \textbf{Ru$\rightarrow$En} & \textbf{En$\rightarrow$Fr} & \textbf{Fr$\rightarrow$En} \\ 
\midrule
Baseline    & 83.8           & 83.9           & 84.9           & 87.0           \\ \midrule
KFMT        & 85.5           & 84.7           & 85.8           & 87.6           \\ 
\hdashline
\textit{~w/o} Sum.    &    84.8       &    84.4    &   85.5           &   87.4         \\ 
\textit{~w/o }Enti.   &    85.0       &    84.5     &   85.5             &    87.5            \\ 
\bottomrule
\end{tabular}
\caption{Ablation study for the summarization and entity translation fusion. \textit{w/o} Sum. and \textit{w/o} Ent. denote that we remove the translation {\small$\mathcal{Y}^s$} and {\small$\mathcal{Y}^b$} when generating the final translation via Eq. \ref{equ:scoring}, respectively.}
\label{tbl:ablation}
\end{table}

\paragraph{Effect of Each Knowledge.} Although the results presented in Table~\ref{tbl:comet_res} highlight the overall effectiveness of our approach, the specific impact of each type of knowledge on the final translation is not entirely clear. To address this, we analyze the influence of different types of knowledge from two perspectives. First, we visualize the proportion of sentences translated with entity ({\small$\mathcal{Y}^e$}), summarization ({\small$\mathcal{Y}^s$}), or without any additional knowledge ({\small$\mathcal{Y}^b$}) across all sentences in the test set, as shown in Figure \ref{fig:propotion}.\footnote{From Section \ref{sec:multi-fusion}, all sentences in our final translation, i.e., {\small$\mathcal{Y}^f$}, are selected from {\small$\mathcal{Y}^b$}, {\small$\mathcal{Y}^e$} or {\small$\mathcal{Y}^s$} by Eq. \ref{equ:scoring}.}
Our observations reveal that, on average, 50\% of the translations come from the baseline system\ignore{, which translates without additional knowledge}. The remaining 50\% of sentences benefit from incorporating summarization or entity translation knowledge, with summarization knowledge being utilized more frequently than entity translation knowledge. This indicates that integrating more relevant knowledge can significantly enhance translation quality, while incorporating redundant knowledge might actually impair it. Second, we perform an ablation study by individually removing either summarization or entity translation knowledge from the knowledge fusion. As shown in Table~\ref{tbl:ablation}, the results tell that summarization knowledge plays a more critical role than entity translation knowledge in enhancing translation performance. For more details of comparing \texttt{SuMT} (or \texttt{EnMT}) with \texttt{Baseline}, please refer to Appendix~\ref{apdx:baseline_sumt_enmt}.

\begin{table}[!t]
\centering
\small
\begin{tabular}{lrrrr}
\toprule
\textbf{System}  & \textbf{En$\rightarrow$Ru} & \textbf{Ru$\rightarrow$En} & \textbf{En$\rightarrow$Fr} & \textbf{Fr$\rightarrow$En} \\ 
\midrule
\multicolumn{5}{c}{\texttt{Perplexity}} \\
\midrule
Baseline & 10.7  & 30.8 & 62.7 & 29.9 \\ 
KFMT     & \bf 7.5  & \bf 29.5 & \bf 57.5 & \bf 29.6 \\ 
\midrule
\multicolumn{5}{c}{\texttt{Coherence}} \\
\midrule
Baseline & 56.6  & 41.0 & 91.7 & 42.1 \\ 
KFMT     & 56.6  & 41.1 & \bf 92.6 & 42.2 \\ 
\bottomrule
\end{tabular}
\caption{Translation fluency evaluation in perplexity and coherence.}
\label{tbl:ppl-coh}
\end{table}

\paragraph{Performance in Translation Fluency.} Since \textit{KFMT} selects translations from multiple translation systems, we investigate whether this affects translation fluency. Fluency refers to how well the translated text aligns with the norms and naturalness of the target language, which in turn enhances its readability and ease of understanding. Building on previous studies \cite{li-etal-2022-contrastive,kallini-etal-2024-mission}, we evaluate translation fluency using two metrics: perplexity and coherence. Specifically, we compute perplexity scores using the \texttt{GPT-2}~\cite{radford_etal_2019_language}, and assess coherence by measuring the similarity between neighboring sentences using the \texttt{SimCSE} \cite{gao-etal-2021-simcse}\ignore{ to compute embeddings. Therefore, the coherence score is defined as: $\textbf{COH} (x_i,x_{i+1})=\frac{\texttt{SimCSE}(x_i)\cdot \texttt{SimCSE} (x_{i+1}) }{||\texttt{SimCSE}(x_i)||\cdot ||\texttt{SimCSE} (x_{i+1})||}$, where \texttt{SimCSE(x)} denotes the pre-trained SimCSE sentence embedding}. 

As shown in Table \ref{tbl:ppl-coh}, \texttt{KFMT} achieves an improvement of 2.5 scores in perplexity over \texttt{Baseline} while it maintains stability and consistency with the \texttt{Baseline} in terms of the coherence metric. while maintaining stability and consistency in coherence scores. This suggests that, despite selecting translations from different systems, our multi-knowledge fusion approach either preserves or enhances the fluency of document-level machine translations. 

\paragraph{GPT-based and Human Evaluations.}

In addition to the automatic evaluation metrics, we employ GPT-based and human evaluation to achieve a more comprehensive assessment of our results.

\begin{table}[!t]
\centering
\small
\begin{tabular}{lcccc}
\toprule
\textbf{System}   & \textbf{En$\rightarrow$Ru} & \textbf{Ru$\rightarrow$En} & \textbf{En$\rightarrow$Fr} & \textbf{Fr$\rightarrow$En} \\ 
\midrule
{Baseline} & 70.1          & 82.7          & 79.7          & 84.0            \\ 
{KFMT}     & \bf 74.6          & \bf 84.7          & \bf 82.9          & \bf 86.1            \\
\bottomrule
\end{tabular}
\caption{Averaged evaluation score of \texttt{GPT-4o}.}
\label{tbl:gpt_eval_table}
\end{table}

Research by \citet{kocmi-etal-2023-large} suggests the superiority of GPT-based evaluations over traditional automatic metrics like BLEU and COMET, with reliable evaluation requiring models outperforming \texttt{GPT-3.5-turbo}. Consequently, we utilized \texttt{GPT-4o} for our evaluations. \texttt{GPT-4o} not only outperforms \texttt{GPT-3.5-turbo} but is also comparable to \texttt{GPT-4}. Specifically, we adopt the prompt template from \citet{kocmi-etal-2023-large} but refined it to address issues with the original prompt, which included unnecessary explanations. This refinement resulted in a stable and consistent format with 100\% reliability. In our prompt, the model is instructed to assign a score from 0 to 100 to the translation results, where 0 indicates ``no retained meaning'' and 100 denotes ``perfect meaning and grammar.'' For the specific prompt used, see Table \ref{table:gpt_eval_prompt} in Appendix~\ref{apdx:prompt_gpt}. Table \ref{tbl:gpt_eval_table} presents the average scores, demonstrating that \texttt{KFMT} outperforms \texttt{Baseline} across all language pairs. 

\begin{figure}[!t]
\centering
\resizebox{0.99\columnwidth}{!}{
\includegraphics[trim={0cm 0cm 0cm 0cm}]{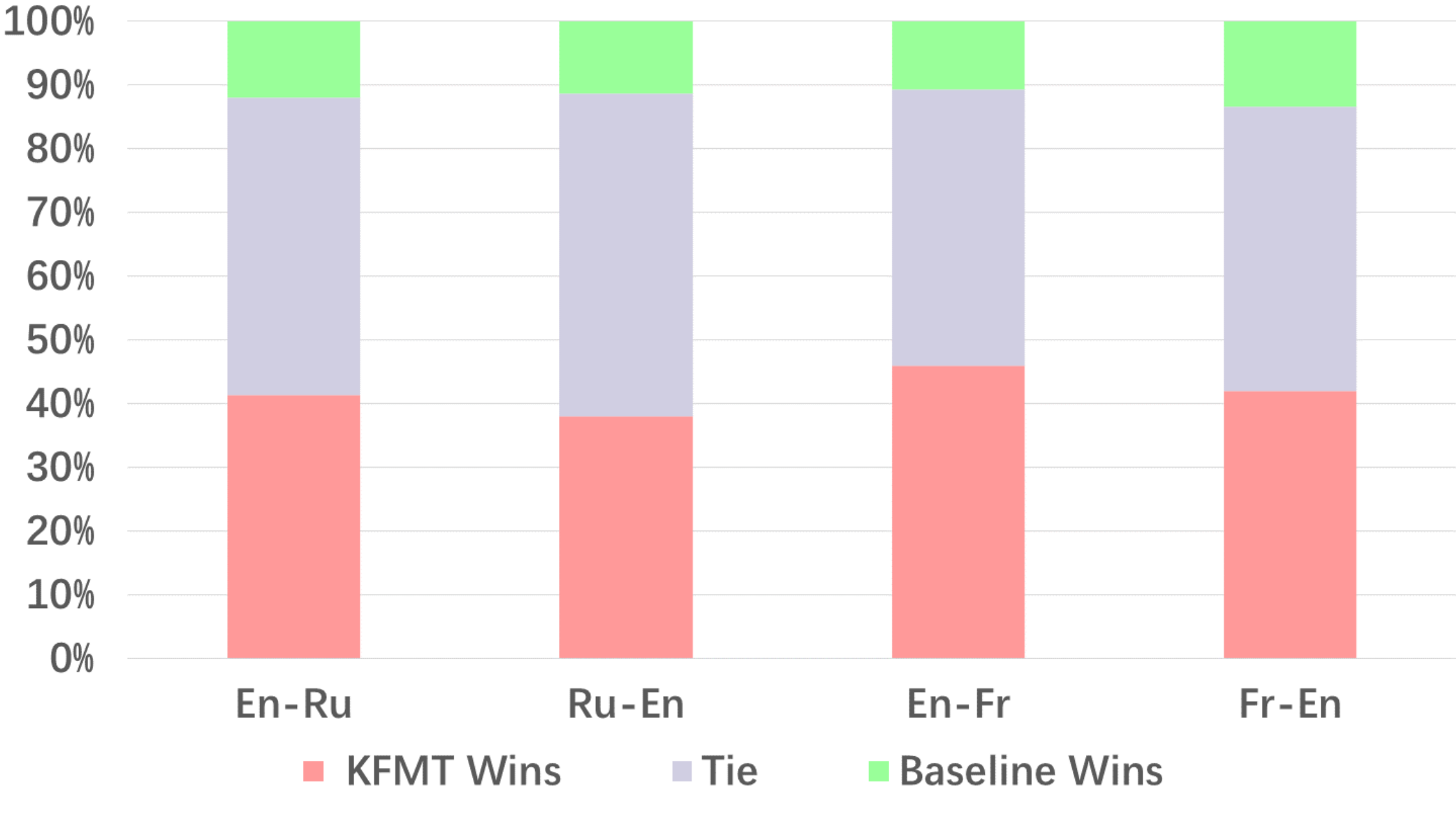}}
\caption{Human evaluation results on the test set when comparing \texttt{KFMT} with \texttt{Baseline}.
}
\label{fig:human}
\end{figure}

We also conduct a human evaluation on the test set. For each document, annotators receive the source document along with translations of \texttt{KFMT} and \texttt{Baseline}, presented in random order. In line with \citet{lyu-etal-2021-encouraging}, annotators are asked to choose from one of three options based on fluency and correctness: (1) the first translation is better, (2) the second translation is better, or (3) both translations are of equal quality. Two annotators, one for En$\rightleftharpoons$Ru and one for En$\rightleftharpoons$Fr, are encouraged to select one of the first two options if they can identify a clear preference, rather than opting for the third. Figure \ref{fig:human} displays the \ignore{human evaluation }results. On average, annotators rate 46.3\% of the cases as having equal quality, while \texttt{KFMT} is preferred over \texttt{Baseline} in 41.8\% of the cases compared to 11.9\%, indicating a clear preference for our approach.

\begin{table}[!t]
\centering
\small
\begin{tabular}{lcccc}
\toprule
               & \textbf{En$\rightarrow$Ru} & \textbf{Ru$\rightarrow$En} & \textbf{En$\rightarrow$Fr} & \textbf{Fr$\rightarrow$En} \\ 
\midrule
Baseline       & 83.8 & 83.9 & 84.9 & 87.0 \\ 
TFMT           & 84.0 & 84.2 & 85.1 & 87.2 \\ 
KFMT           & \bf 85.5 & \bf 84.7 & \bf 85.8 & \bf 87.6 \\ 
\bottomrule
\end{tabular}
\caption{Performance comparison of token-level knowledge fusion (\texttt{TFMT}) and our \texttt{KFMT}.}
\label{tbl:token_fusion}
\end{table}

\paragraph{Comparison to Token-Level Knowledge Fusion.} 
 Our approach to multi-knowledge fusion operates at the sentence level, as it involves reranking translations on a per-sentence basis. In contrast, we investigate token-level multi-knowledge fusion. For a decoder-only model used in translation, the probability of token $t$ at the $i$-th time step is given by:
\begin{equation}
\small
P_{}(t_i) = P_{}(t_i \mid \mathcal{P}, \mathcal{X}, t_{j < i}),
\end{equation}
where $\mathcal{P}$ represents the prompt, $\mathcal{X}$ denotes the source document, and $t_{j < i}$ indicates the previously generated tokens. Let $P_{B}(t_i)$,$P_{S}(t_i)$,$P_{E}(t_i)$ represent the probabilities of token at the $i^{\text{th}}$ time step for the systems of \texttt{Baseline}, \texttt{SuMT}, and \texttt{EnMT}, respectively. Motivated by~\citet{hoang-etal-2024-fly}, we perform token-level fusion by combining these systems in an ensemble. We assign weight parameters $\lambda_1$, $\lambda_2$ and $\lambda_3$ to the respective systems, ensuring that their sum equals 1 ($\lambda_1 + \lambda_2 + \lambda_3  = 1$). Thus, the probability of $i$-th token $t_i$\ignore{ based on token-level fusion} is:
\begin{equation}
\centering
\small
P_{\text{ensemble}}(t_i) = \lambda_1 P_{B}(t_i) + \lambda_2 P_{S}(t_i) + \lambda_3 P_{E}(t_i).
\end{equation}

During inference, we set the temperature to 0 and the weight to $\lambda_1=0.4$, $\lambda_2=0.3$, and $\lambda_3=0.3$, respectively. Table \ref{tbl:token_fusion} compares the performance. It shows that token-level knowledge fusion (i.e., \texttt{TFMT}) provides an average improvement of 0.2 COMET scores over \texttt{Baseline}. However, it performs less effectively compared to our proposed \texttt{KFMT}, which achieves an average improvement of 0.7 COMET scores.

\section{Related Work}
\paragraph{Conventional Document-Level Machine Translation.} Conventional DMT, which are built upon the Transformer~\cite{vaswani-etal-2017-transformer}, have made significant advancements in recent years. These models generally fall into two main categories. The first category focuses on translating document sentences one by one while incorporating document-level context \cite{zhang-etal-2018-improving,maruf-etal-2019-selective,zheng-etal-2020-towards}. The second category extends the translation unit from a single sentence to multiple sentences \cite{tiedemann-etal-2017-neural,agrawal-etal-2018-contextual,zhang-etal-2020-longshort} or the entire document \cite{junczys-etal-2019-microsoft,liu-etal-2020-multilingual,bao-etal-2021-gtransformer,li-etal-2023-ptransfomer}.

\paragraph{LLMs for Document-Level Machine Translation.} 

The adaptation of LLMs for DMT is an emerging research area with significant potential. Current research in this domain primarily explores two types of approaches: supervised fine-tuning and prompt engineering. Supervised fine-tuning aims to enhance LLMs' capabilities for document-level machine translation through targeted training. For example, \citet{zhang-etal-2023-machine} fine-tune the model using the Q-LORA method and compare its document translation performance with the prompt engineering approach. \citet{wu-etal-2024-adapting} introduce a two-step fine-tuning method. Initially, LLMs are fine-tuned on monolingual data, and subsequently, they are further fine-tuned on parallel documents. \citet{li-etal-2024-enhancing} propose a hybrid approach that integrates sentence-level fine-tuning instructions into the document-level fine-tuning process, which aims to improve overall translation performance. Furthermore, \citet{lyu-etal-2024-dempt} show that LLMs can be more effectively adapted for context-aware NMT by discriminately modeling and utilizing both inter- and intra-sentence contexts. 

Prompt engineering focuses on designing prompts to optimize DMT. \citet{wang-etal-2023-documentlevel} investigate how various document translation prompts impact translation performance and assess the capabilities of different LLMs. Additionally, \citet{cui-etal-2024-cfficiently} explore the use of contextual summaries to select the most relevant examples, thus enhancing sentence context for translation. Moreover, \citet{wang-etal-2024-delta} propose a document-level translation agent that improves the consistency and accuracy of document translation by utilizing a multi-level memory structure. Our work is aligned with prompt engineering but diverges from previous approaches by proposing the integration of various types of knowledge to enhance document translation. The most relevant work to ours is \citet{he-etal-2024-exploring} which investigates the use of knowledge for sentence-level translation. In contrast, our approach extends this research to DMT by incorporating document-level knowledge.

\section{Conclusion}
In this paper, we propose a multi-knowledge fusion approach that mimics human translators for document-level machine translation. Our approach explicitly combines different types of knowledge to enhance translation quality. It involves three key steps: First, it acquires two types of document-level knowledge—summarization and entity translation. Next, it integrates this knowledge to improve the translation process. Finally, recognizing that different knowledge sources may impact sentence translation differently, we optimize results using a multi-knowledge fusion strategy for refinement and ranking. Experiments across eight DMT tasks demonstrate that our approach consistently enhances performance across three different LLMs.


\section*{Limitations}
Our approach has only been validated on a news dataset, and all its language pairs include English, so its effectiveness on a broader range of datasets and non-English language pairs remains uncertain. Moreover, in terms of performance improvements across the three LLMs, our approach demonstrates more significant gains on LLMs with weaker document translation performance, whereas the relative improvement is less pronounced on LLMs with stronger translation capabilities.

\bibliography{custom}

\appendix

\section{Data Statistics}
\label{apdx:data_stat}

Table~\ref{tbl:data_stat} presents the detailed statistics of test sets. On average, each document across the four language pairs contains between 37 and 39 sentences.

\begin{table}[!ht]
\centering
\begin{tabular}{lcc}
\toprule
\textbf{Dataset} & \textbf{\# Document}  & \textbf{\# Sentence} \\ \midrule
{De$\rightleftharpoons$En}   & 150    &   5,967        \\ 
{Es$\rightleftharpoons$En}   & 150     &   5,815       \\ 
{Ru$\rightleftharpoons$En}   & 150    &   5,794        \\ 
{Fr$\rightleftharpoons$En}   & 150     &   5,619       \\ \bottomrule
\end{tabular}
\caption{Data statistics on our test sets.}
\label{tbl:data_stat}
\end{table}

\section{Performance in dCOMET and BLEU}
\label{apdx:bleuAndDComet}
Unlike traditional sentence-level COMET, the document-level COMET (dCOMET) introduced by \citet{giorgos-etal-2022-embarrassingly} takes into account the context of previous sentences during encoding. This approach results in more accurate evaluations of document-level translations. Following the work of ~\cite{wang-etal-2024-delta}, we employed a reference-free model to obtain dCOMET scores. However, we replaced the original model, \texttt{wmt21-comet-qe-mqm}, with the latest version, \texttt{wmt22-cometkiwi-da}.The dCOMET scores, derived using the \texttt{wmt22-cometkiwi-da} model, are presented in Table~\ref{tbl:dCOMET}. As shown, the performance trend of dCOMET closely follows that of sentence-level COMET, which is also presented in Table~\ref{tbl:dCOMET}.

Table~\ref{tbl:bleu} presents the detailed performance in sentence-level BLEU~\cite{papineni-etal-2002-bleu}. From it, we can observe that our approach achieves consistent and stable improvements in BLEU scores, although the gains in both COMET and dCOMET scores are relatively modest. 

\begin{table*}[th]
\centering
\small
\begin{tabular}{lccccccccc}
\toprule
\textbf{System} & \textbf{En$\rightarrow$De} & \textbf{De$\rightarrow$En} & \textbf{En$\rightarrow$Es} & \textbf{Es$\rightarrow$En} & \textbf{En$\rightarrow$Ru} & \textbf{Ru$\rightarrow$En} & \textbf{En$\rightarrow$Fr} & \textbf{Fr$\rightarrow$En} & \textbf{Average}\\ 
\midrule
\multicolumn{9}{c}{\texttt{LLaMA3-8B-Instruct}} \\ 
\midrule
Baseline     & 83.8 & 82.4 & 85.3 & 84.3 & 81.8 & 80.8 & 85.3 & 83.0 & 83.3\\
Reranking     & 84.0 & 82.4 & 85.4 & 84.3 & 82.2 & 81.0 & 85.4 & 83.0 &83.5 \\
\hdashline
SuMT     & 83.8 & 82.4 & 85.3 & 84.3 & 81.8 & 80.8 & 85.3 & 83.0 & 83.3 \\
EnMT     & 83.8 & 82.4 & 85.3 & 84.2 & 81.7 & 80.8 & 85.3 & 83.0 & 83.3\\ \hdashline
\(\text{KFMT}\)     &\textbf{84.3} &\textbf{82.7} &\textbf{85.8} &\textbf{84.6} &\textbf{82.8} &\textbf{81.4} &\textbf{85.8} &\textbf{83.5} &\textbf{83.9}\\
\midrule
\multicolumn{9}{c}{\texttt{Mistral-Nemo-Instruct}} \\ 
\midrule
Baseline     & 83.7 & 82.6 & 84.8 & 84.4 & 83.0 & 81.1 & 85.3 & 83.4 & 83.5\\ 
Reranking     & 84.0 & 82.6 & 85.0 & 84.4 & 83.2 & 81.1 & 85.5 & 83.4 & 83.6\\
\hdashline
SuMT    & 83.7 & 82.4 & 84.9 & 84.4 & 83.1 & 80.9 & 85.4 & 83.2 & 83.5 \\ 
EnMT    & 83.8 & 82.6 & 84.9 & 84.5 & 83.1 & 81.1 & 85.5 & 83.3 & 83.6 \\ \hdashline
\(\text{KFMT}\)     &\textbf{84.4} &\textbf{82.7} &\textbf{85.7} &\textbf{84.7} &\textbf{83.5} &\textbf{81.3} &\textbf{85.8} &\textbf{83.5} &\textbf{84.0}\\
\midrule
\multicolumn{9}{c}{\texttt{GPT-4o-mini}} \\ 
\midrule
Baseline     & 84.6 & 82.7 & 86.0 & 84.6 & 83.6 & 81.3 & 85.9 & 83.5 & 84.0\\
Reranking     & 84.7 & 82.7 & 86.0 & 84.6 & 83.6 & 81.3 & 85.9 & 83.6 & 84.0\\
\hdashline
SuMT     & 84.6 & 82.7 & 86.0 & 84.6 & 83.6 & 81.3 & 85.9 & 83.5 & 84.0\\ 
EnMT     & 84.6 & 82.7 & 86.0 & 84.6 & 83.5 & 81.2 & 85.8 & 83.4 & 84.0\\ \hdashline
\(\text{KFMT}\)    &\textbf{84.8} &\textbf{82.9} &\textbf{86.2} &\textbf{84.8} &\textbf{83.9} &\textbf{81.5} &\textbf{86.1} &\textbf{83.7} &\textbf{84.2}\\
\bottomrule
\end{tabular}
\caption{Performance in document-level (dCOMET) score.}
\label{tbl:dCOMET}
\end{table*}

\begin{table*}[th]
\centering
\small
\begin{tabular}{lccccccccc}
\toprule
\textbf{System} & \textbf{En$\rightarrow$De} & \textbf{De$\rightarrow$En} & \textbf{En$\rightarrow$Es} & \textbf{Es$\rightarrow$En} & \textbf{En$\rightarrow$Ru} & \textbf{Ru$\rightarrow$En} & \textbf{En$\rightarrow$Fr} & \textbf{Fr$\rightarrow$En} & \textbf{Average}\\ 
\midrule
\multicolumn{9}{c}{\texttt{LLaMA3-8B-Instruct}} \\ 
\midrule
Baseline     & 31.2 & 39.5 & 42.0 & 45.4 & 27.7 & 31.8 & 35.9 & 37.6 &36.4\\ 
Reranking     & 31.6 & 39.5 & 42.2 & 45.4 & 28.1 & 32.3 & 36.1 & 37.6 &36.6\\
\hdashline
SuMT     & 31.0 & 40.0 & 42.0 & 45.7 & 27.4 & 32.4 & 35.7 & 38.4 &36.6\\
EnMT     & 31.0 & 40.0 & 41.6 & 44.0 & 27.3 & 31.2 & 35.6 & 36.4 &35.9\\ \hdashline
\(\text{KFMT}\)     & \textbf{32.1} & \textbf{40.4} & \textbf{42.5} & \textbf{46.1} & \textbf{28.8} & \textbf{32.9} & \textbf{36.5} & \textbf{38.4} &\textbf{37.2}\\
\midrule
\multicolumn{9}{c}{\texttt{Mistral-Nemo-Instruct}} \\ 
\midrule
Baseline     & 34.6 & 42.6 & 43.3 & 48.0 & 31.7 & 35.7 & 38.2 & 40.2 &39.3\\
Rerankding     & 35.0 & 42.7 & 43.5 & 48.0 & 32.0 & 35.8 & 38.5 & 40.2 &39.5\\
\hdashline
SuMT     & 34.2 & 39.5 & 43.2 & 45.4 & 31.2 & 33.4 & 38.5 & 37.8 &37.9\\ 
EnMT     & 34.5 & 41.3 & 43.1 & 47.0 & 31.4 & 34.8 & 38.3 & 39.3 &38.7\\ \hdashline
\(\text{KFMT}\)     & \textbf{35.8} & \textbf{43.2} & \textbf{44.6} & \textbf{48.5} & \textbf{32.5} & \textbf{36.1} & \textbf{39.1} & \textbf{40.7} &\textbf{40.1}\\
\midrule
\multicolumn{9}{c}{\texttt{GPT-4o-mini}} \\ 
\midrule
Baseline     & 41.2 & 44.0 & 48.6 & 50.1 & 36.0 & 37.0 & 42.0 & 41.4 &42.5\\
Reranking     & 41.3 & 44.0 & 48.6 & 50.1 & 36.0 & 37.1 & 42.0 & 41.5 &42.6\\
\hdashline
SuMT     & 41.0 & 44.0 & 48.4 & 50.0 & 35.8 & 36.9 & 41.9 & 41.4 &42.4\\ 
EnMT     & 39.9 & 42.7 & 47.5 & 48.9 & 34.8 & 35.8 & 41.1 & 40.2 &41.4\\ \hdashline
\(\text{KFMT}\)     & \textbf{41.5} & \textbf{44.2} & \textbf{49.0} & \textbf{50.4} & \textbf{36.5} & \textbf{37.3} & \textbf{42.4} & \textbf{41.7} &\textbf{42.9}\\
\bottomrule
\end{tabular}
\caption{Performance in SacreBLEU score.}
\label{tbl:bleu}
\end{table*}

\section{Performance in LTCR}
\label{apdx:ltcr}
Following \citet{lyu-etal-2021-encouraging}, we use LTCR to assess lexical translation consistency. Table~\ref{tbl:ltcr} compares the performance of \texttt{LLaMA3-8B-Instruct} in the LTCR metric. The results show that our approach improves lexical consistency with an average gain of 1.62 points on the LTCR score.

\begin{table*}[t]
\centering
\small
\begin{tabular}{lccccccccc}
\toprule
\textbf{System} & \textbf{En$\rightarrow$De} & \textbf{De$\rightarrow$En} & \textbf{En$\rightarrow$Es} & \textbf{Es$\rightarrow$En} & \textbf{En$\rightarrow$Ru} & \textbf{Ru$\rightarrow$En} & \textbf{En$\rightarrow$Fr} & \textbf{Fr$\rightarrow$En} & \textbf{Average}\\ 
\midrule
Baseline & 86.67 & 81.25 & 74.19 & 85.71 & 42.42 & 66.67 & 83.33 & 63.64 &73.00\\
KFMT & \bf 93.33 & \bf 82.35 & \bf 75.00 & \bf 86.67 & \bf 45.45 & 66.67 & 83.33 & \bf 64.12 & \bf 74.62\\
\bottomrule
\end{tabular}
\caption{Lexical translation consistency evaluation in LTCR.}
\label{tbl:ltcr}
\end{table*}

\section{Performance in BlonDe}
\label{apdx:blonde}

Table~\ref{tbl:blonde} compares the performance of \texttt{LLaMA3-8B-Instruct} in BlonDe. It shows that our approach outperforms \texttt{Baseline} with 4.6 BlonDe scores. 

\begin{table*}[!th]
\centering
\small
\begin{tabular}{lccccc}
\toprule
\textbf{System}   & \textbf{De$\rightarrow$En} & \textbf{Es$\rightarrow$En} & \textbf{Ru$\rightarrow$En} & \textbf{Fr$\rightarrow$En} & \textbf{Average}\\ 
\midrule
{Baseline} &  49.4         & 59.4          &     41.1      &     50.1   &  50.0   \\ 
{KFMT}     & \bf  58.9         & \bf  60.5         & \bf   46.8        & \bf  52.0    & \bf  54.6       \\
\bottomrule
\end{tabular}
\caption{Performance in BlonDe score.}
\label{tbl:blonde}
\end{table*}

\section{Comparison of \texttt{Baseline}, \texttt{SuMT}, and \texttt{EnMT}}
\label{apdx:baseline_sumt_enmt}
\begin{figure}[!t]
\centering
\resizebox{0.99\columnwidth}{!}{
\includegraphics[trim={0cm 0cm 0cm 0cm}]{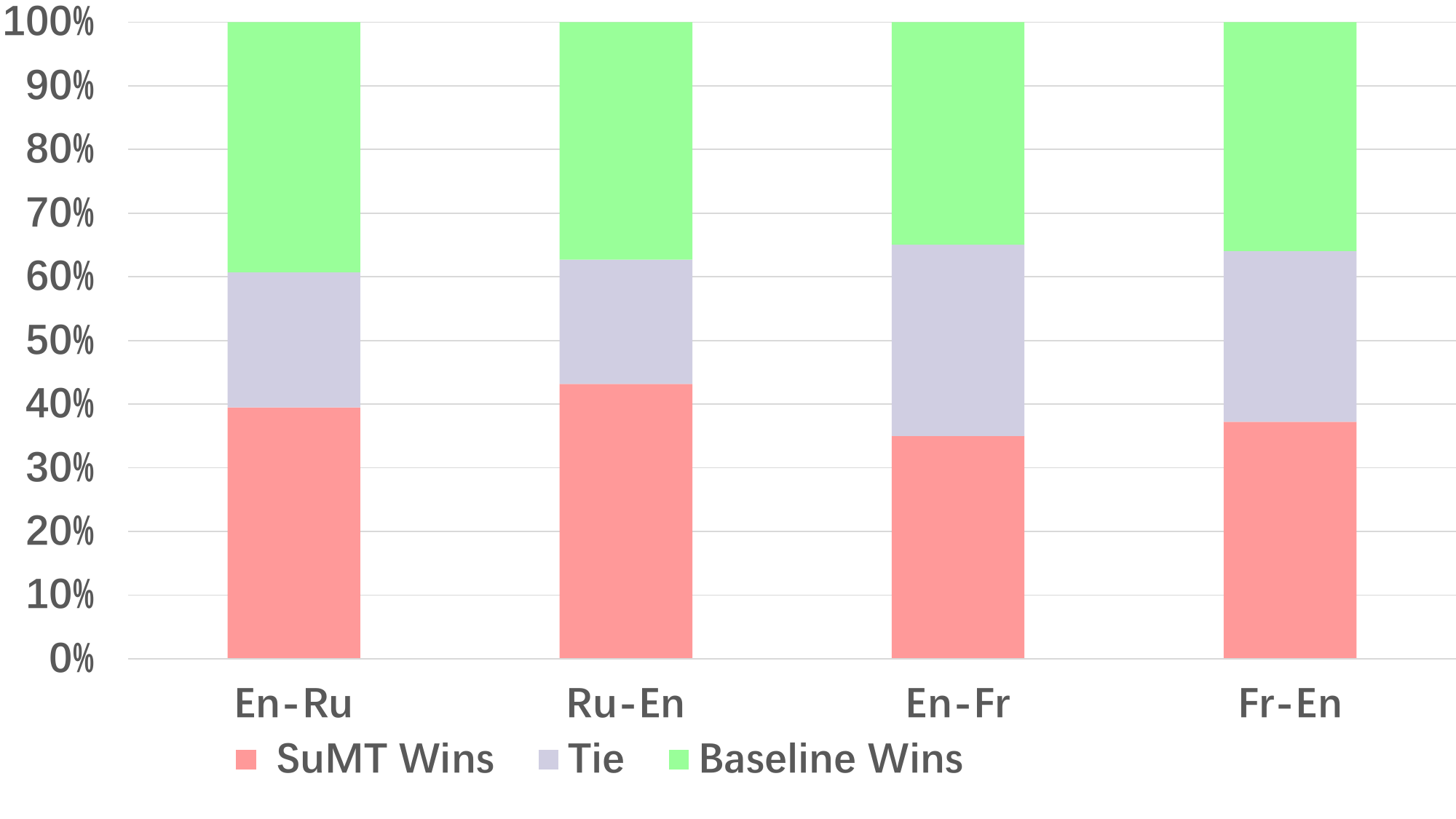}}
\caption{Comparison of \texttt{SuMT} and the \texttt{Baseline} in terms of sentence-level reference-based COMET scores.
}
\label{fig:SuMT_vs_Base}
\end{figure}

\begin{figure}[!t]
\centering
\resizebox{0.99\columnwidth}{!}{
\includegraphics[trim={0cm 0cm 0cm 0cm}]{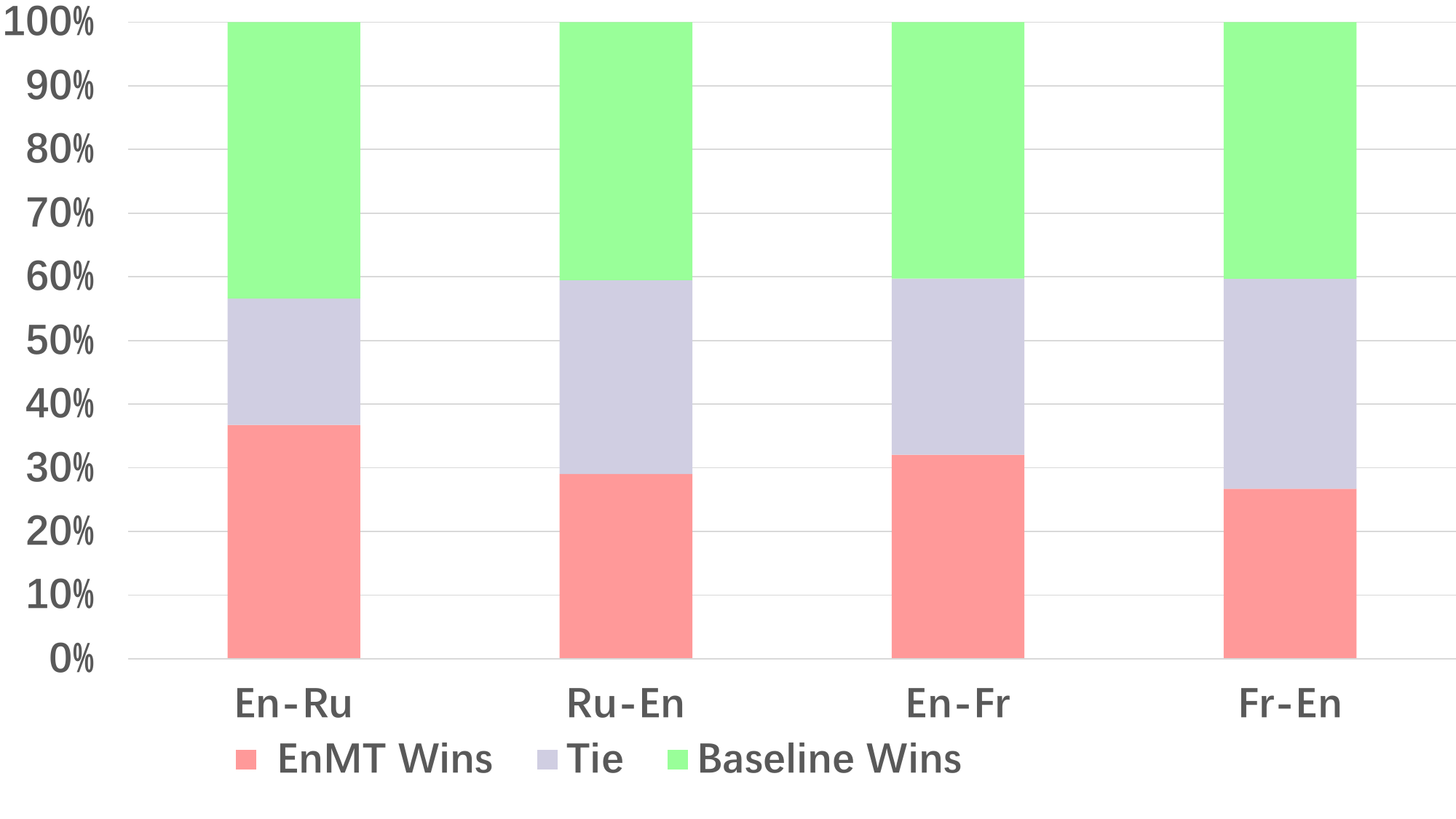}}
\caption{Comparison of \texttt{EnMT} and the \texttt{Baseline} in terms of sentence-level reference-based COMET scores.
}
\label{fig:EnMT_vs_Base}
\end{figure}

Incorporating a single source of knowledge benefits only certain sentences in the translation, while others may be negatively affected. Figure~\ref{fig:SuMT_vs_Base} and Figure~\ref{fig:EnMT_vs_Base} present the results of \texttt{LLaMA3-8B-Instruct} on the test set, comparing \texttt{SuMT} (or \texttt{EnMT}) with the \texttt{Baseline} at sentence-level reference-based COMET score by \texttt{wmt22-comet-da}. Note that we consider two translations to be tied if their COMET scores differ by no more than 0.08.

\section{Performance of Summarization and Entity Translation}
\label{apdx:summarziation_entity}
While recent studies have demonstrated that LLMs are highly effective in generating summaries~\cite{pu-etal-2023-summarization,tianyi-etal-2024-benchmarking} and translating entities~\cite{he-etal-2024-exploring}, their performance on our experimental datasets remains uncertain. 

The studies by \citet{jiaan-etal-2023-nlgEval} demonstrate that GPT evaluations achieve remarkable performance and closely align with human assessments in various NLP tasks. Therefore, we adopt and modify the reference-free scoring prompt from \citet{jiaan-etal-2023-nlgEval} to evaluate the quality of our summarization knowledge and entity translation knowledge. Table~\ref{tbl:prompt_eval_sum_and_enti} presents the prompts used in our evaluation of the generated summarization knowledge and entity translation knowledge with \texttt{GPT-4o}. Table~\ref{tbl:gpt_eval_for_sum_enti} provides the corresponding evaluation scores. It shows that our summarization and entity translation are of good quality.

\begin{table*}[t]
\centering
\small
\resizebox{\textwidth}{!}{
\begin{tabular}{lp{0.15\textwidth}p{0.75\textwidth}}
\toprule
\textbf{ID} &\textbf{Task} & \textbf{Prompt Template} \\
\midrule
\#1 & Summarization &  Score the following summarization for overall quality on a continuous scale from 0 to 100. A score of zero means "poor quality" (disjointed, hard to read, or containing significant factual inaccuracies), and a score of one hundred means "excellent quality" (fluent, coherent, and consistent with the key ideas of the original text).\newline \newline
Overall Quality measures both:\newline
1. Fluency: Whether the summarization is well-written, grammatically correct, and easy to understand, with smooth sentence transitions and a natural flow.\newline
2. Consistency: Whether the summarization accurately conveys the main points, key details, and intended meaning of the original text, without introducing errors, distortions, or irrelevant information. \newline \newline
Original Text: <Original text> \newline 
Summarization: <Summarization> \newline 
Score:\\
\midrule
\#2 & Entity Translation & Score the following \textbf{\textit{<src\_lang>}} translation of extracted entities from original text for overall quality on a continuous scale from 0 to 100. A score of zero means "poor quality" (missing important entities or containing significant translation errors), and a score of one hundred means "excellent quality" (all key entities are extracted and translated accurately). \newline \newline
Overall Quality measures both: \newline 
1. Correctness of Entity Extraction: Whether the extracted entities include all critical persons, locations, organizations, dates, and other relevant information explicitly mentioned in the original text. \newline 
2. Translation Accuracy: Whether the English translations are faithful to the meaning, spelling, and nuances of the original entities. \newline \newline
Original Text: <Original text> \newline 
Extracted Entities (Original): <the list of extracted entities in the original language> \newline 
Extracted Entities (Translated): <the list of extracted entities in \textbf{\textit{<src\_lang>}}> \newline 
Score: \\
\bottomrule
\end{tabular}
}
\caption{Prompt Templates for evaluating summrization and entity translation.}
\label{tbl:prompt_eval_sum_and_enti}
\end{table*}

\begin{table*}[!th]
\centering
\small
\begin{tabular}{lccccc}
\toprule
\textbf{Task}   & \textbf{En$\rightarrow$Ru} & \textbf{Ru$\rightarrow$En} & \textbf{En$\rightarrow$Fr} & \textbf{Fr$\rightarrow$En} & \textbf{Average}\\ 
\midrule
{Summarization} &  78.0         & 78.6          &     77.1      &     80.5   &  78.6   \\ 
{Entity Translation}     &   87.1         &   87.1         &    91.7        &   83.6    &   87.4       \\
\bottomrule
\end{tabular}
\caption{Averaged evaluation score of \texttt{GPT-4o} for Summarization and Entity Translation.}
\label{tbl:gpt_eval_for_sum_enti}
\end{table*}

\section{Prompt for \texttt{GPT} Evaluation}
\label{apdx:prompt_gpt}

Table~\ref{table:gpt_eval_prompt} presents the prompt template used in \texttt{GPT} evaluation in Section~\ref{sec:expr_analysis}. The \textcolor{DeepYellow}{highlighted} part demonstrates our difference from the template described in \citet{kocmi-etal-2023-large}. By specifying that the scores be returned in dictionary format, we achieve 100\% consistency in format output.

\begin{table*}[!t]
\centering
\small
\begin{tabularx}{\textwidth}{X}
\toprule
\textbf{Prompt Template} \\ 
\midrule
Score the following translation from \textbf{\textit{<src\_lang>}} to \textbf{\textit{<tgt\_lang>}} with respect to the human reference on a continuous scale from 0 to 100, where a score of zero means "no meaning preserved" and a score of one hundred means "perfect meaning and grammar". \textcolor{DeepYellow}{Please return the score you gave in the dictionary format of \{\{"score": score\}\}. You only need to give the score, no additional explanation is needed.}\\
\textbf{\textit{<src\_lang>}} source: \textbf{\textit{<src\_seg>}} \\
\textbf{\textit{<tgt\_lang>}} human reference: \textbf{\textit{<ref\_seg>}} \\
\textbf{\textit{<tgt\_lang>}} translation: \textbf{\textit{<tgt\_seg>}} \\
Score: \\
\bottomrule
\end{tabularx}
\caption{Prompt template used in \texttt{GPT} evaluation.}
\label{table:gpt_eval_prompt}
\end{table*}

\section{Prompt for formatting LLM's answer}
\label{apdx:prompt_format}
Table~\ref{tbl:format_prompt} presents the prompt we used for formatting LLM's answer, including summarization, entity translation and document translation.

\begin{table*}[t]
\centering
\small
\begin{tabular}{lp{0.18\textwidth}p{0.75\textwidth}}
\toprule
\textbf{ID} &\textbf{Task} & \textbf{Prompt Template} \\
\midrule
\#1 & Formatting \newline summarization & I hope you can return the summarization in dictionary format. The format of the dictionary is: \{\textbackslash" summarization \textbackslash": your summarization\} \\
\midrule
\#2 &Formatting \newline entity translation & I hope you can return the entity and its translation in dictionary format. The key of the dictionary is the entity, and the value is the translation of the entity. \\
\midrule
\#3 & Formatting \newline document translation & I hope you can return your translation results in dictionary format. The keys of the dictionary should be the sentence numbers, and the values should be the translation results of the sentences. For example, if your text consists of two sentences, the format of your final translation results should be: \{`\#1': translation result of sentence 1, `\#2': translation result of sentence 2\}. \\
\bottomrule
\end{tabular}
\caption{Prompt Templates for formatting LLMs' answer.}
\label{tbl:format_prompt}
\end{table*}

\end{document}